\documentclass[runningheads]{llncs}
\usepackage{graphicx}
\usepackage{amsmath,amssymb} % define this before the line numbering.
\usepackage{color}

\usepackage{times}
\usepackage{epsfig}
\usepackage{subfigure}
\usepackage{multirow}
\usepackage{url}
\usepackage{bm}

\begin{document}
\pagestyle{headings}
\mainmatter

\def\ACCV20SubNumber{389}  % Insert your submission number here

%===========================================================
\title{FC$^2$RN: A Fully Convolutional Corner Refinement Network for Accurate Multi-Oriented\\ Scene Text Detection} % Replace with your title
% SUBMISSION
% \titlerunning{ACCV-20 submission ID \ACCV20SubNumber}
% \authorrunning{ACCV-20 submission ID \ACCV20SubNumber}

% \author{Anonymous ACCV 2020 submission}
% \institute{Paper ID \ACCV20SubNumber}

% CAMERA READY
\titlerunning{FC$^2$RN for Accurate Multi-Oriented Scene Text Detection}
\authorrunning{X. Qin, Y. Zhou et al.}

\author{Xugong Qin, Yu Zhou, Dayan Wu, Yinliang Yue, and Weiping Wang}
\institute{Institute of Information Engineering, Chinese Academy of Sciences, Beijing, China\\
School of Cyber Security, University of Chinese Academy of Sciences, Beijing, China\\
\{qinxugong, zhouyu, wudayan, yueyinliang, wangweiping\}@iie.ac.cn}

\maketitle

%===========================================================
\begin{abstract}
Recent scene text detection works mainly focus on curve text detection.
However, in real applications, the curve texts are more scarce than the multi-oriented ones.
Accurate detection of multi-oriented text with large variations of scales, orientations, and aspect ratios is of great significance.
Among the multi-oriented detection methods, direct regression for the geometry of scene text shares a simple yet powerful pipeline and gets popular in academic and industrial communities,
but it may produce imperfect detections, especially for long texts due to the limitation of the receptive field.
In this work, we aim to improve this while keeping the pipeline simple.
A fully convolutional corner refinement network (FC$^2$RN) is proposed for accurate multi-oriented text detection, in which an initial corner prediction and a refined corner prediction are obtained at one pass.
With a novel quadrilateral RoI convolution operation tailed for multi-oriented scene text,
the initial quadrilateral prediction is encoded into the feature maps which can be further used to predict offset between the initial prediction and the ground-truth as well as output a refined confidence score.
Experimental results on four public datasets including MSRA-TD500, ICDAR2017-RCTW, ICDAR2015, and COCO-Text demonstrate that FC$^2$RN can outperform the state-of-the-art methods.
The ablation study shows the effectiveness of corner refinement and scoring for accurate text localization.
\end{abstract}

%===========================================================

\section{Introduction}

Reading text in the wild has attracted lots of attention due to its wide applications in scene understanding, license plate recognition, autonomous navigation, and document analysis.
As a prerequisite of text recognition, text detection plays an essential role in the whole procedure of scene text understanding.
Despite great progress achieved by recent scene text detection methods inspired by object detection and segmentation methods, detecting text in natural images remains very challenging due to large variations of scales, orientations and aspect ratios as well as low qualities and perspective distortions.
Although curve text detection attracts lots of attention recently, the proportion of curve text in reality is relatively small.
Accurate multi-oriented text detection is still one of the most important open problems to be solved.

\begin{figure}[!htb]
\centering
\subfigure[]{
   \includegraphics[width=0.4\linewidth]{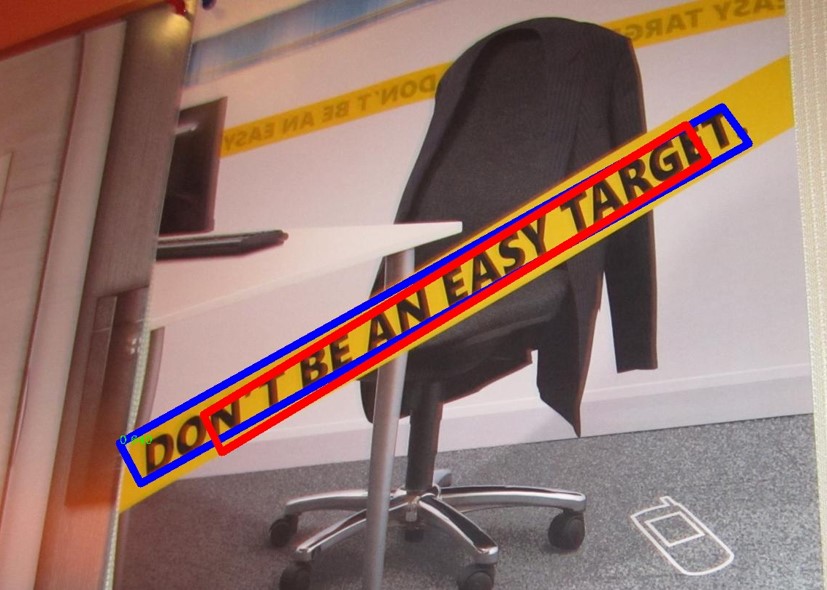}
   \label{fig:1a}
}
\subfigure[]{
   \includegraphics[width=0.4\linewidth]{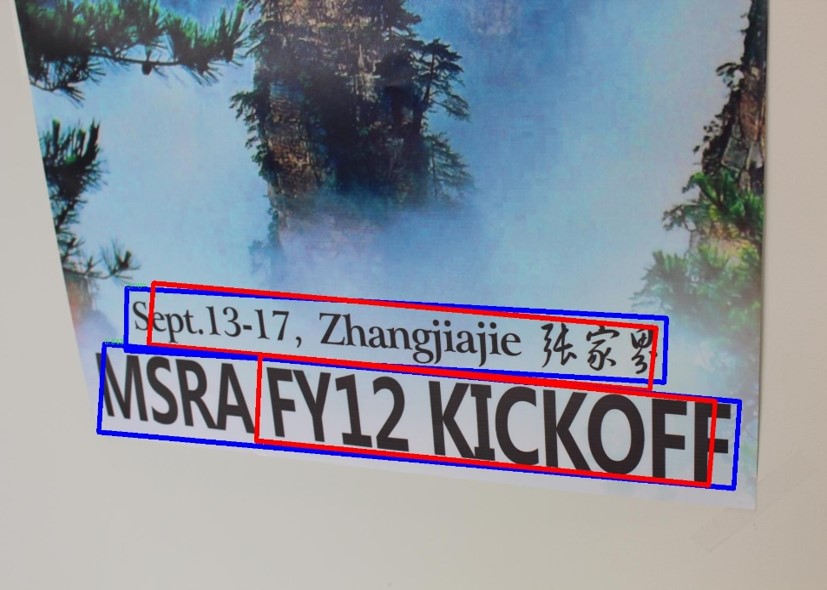}
   \label{fig:1b}
}
\caption{Detection results of direct regression and FC$^2$RN on long text lines. The red and the blue boxes are results of direct regression and FC$^2$RN respectively.}
\label{fig: figure1}
\end{figure}

Recently, various methods \cite{zhou2017east,liu2017deep,he2017deep,liao2018textboxes++,ma2018arbitrary,lyu2018multi,liao2018rotation,li2018shape,zhang2019look,wang2018geometry,liu2019omnidirectional} are proposed to detect multi-oriented text.
All of the methods can be roughly divided into anchor-based methods and anchor-free methods.
Anchor-based methods assume a set of prior boxes for reference which simplifies the problem to learn relative offsets to anchors.
Anchor-free regression methods get rid of complicate design of anchor boxes and directly regress geometry of text, which makes it a clear and effective pipeline for scene text detection \cite{zhou2017east}.
However, due to large variations of scales, orientations and aspect ratios, direct regression may produce unsatisfactory results as shown in Fig. \ref{fig:1a}.
Recent research also reveals performance degradation when training with texts containing large rotation variations \cite{liu2019omnidirectional,xu2019geometry}.
It also shows deficiency when detecting long texts with large aspect ratios because of the limitation of the receptive field as shown in Fig. \ref{fig:1b}.

\begin{figure}[!htb]
\begin{center}
   \includegraphics[width=0.55\linewidth]{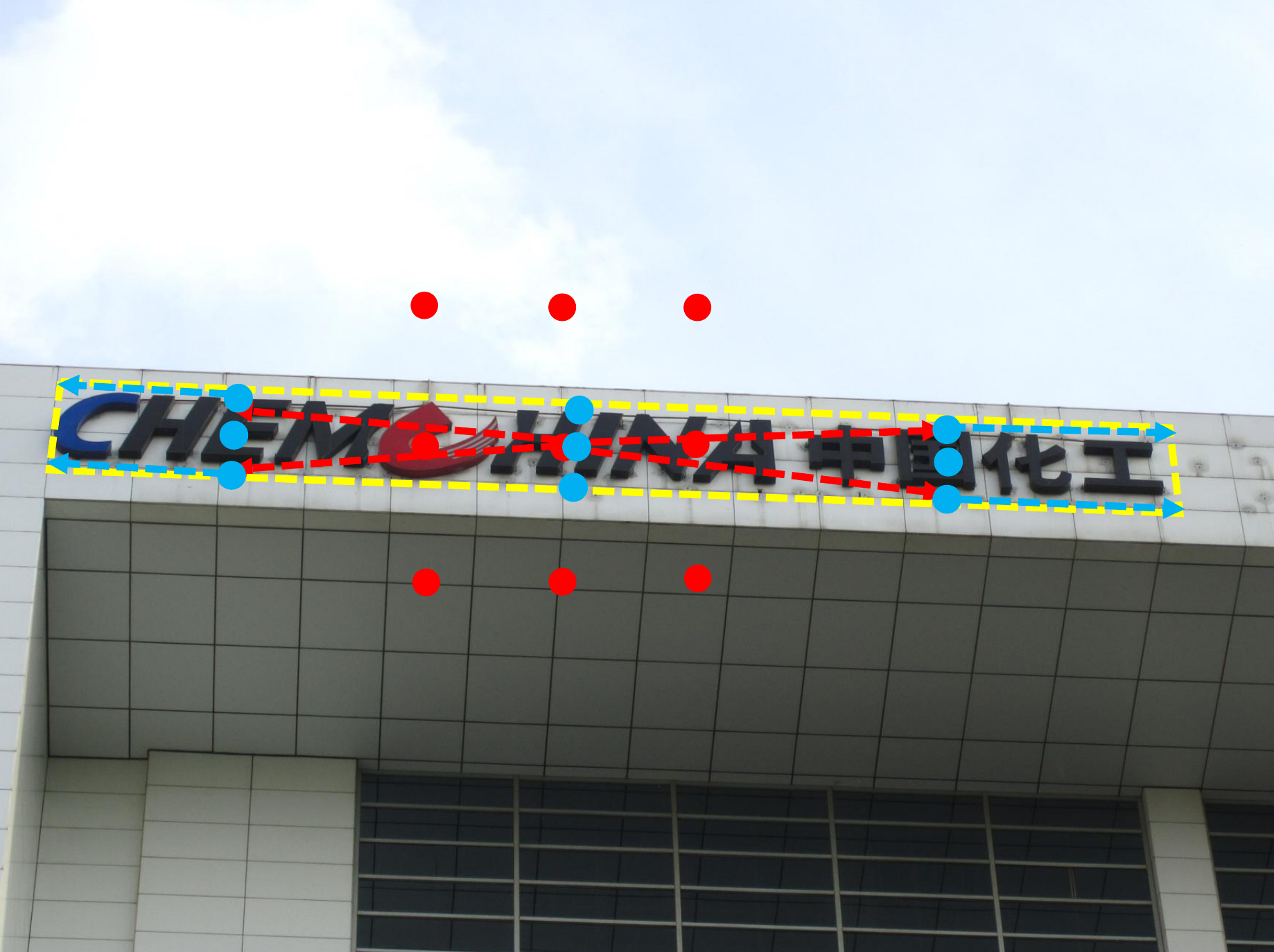}
\end{center}
   \caption{Illustration of our proposed method. The red and blue dashed lines show the process of direct regression and corner refinement. The dashed box in yellow denotes the final detection result. The red and blue solid points represent the sampling positions of standard convolution and QRC.}
\label{fig:illustration}
\end{figure}

To localize multi-oriented text accurately while keeping a simple pipeline, we propose a fully convolutional corner refinement network (FC$^2$RN) for multi-oriented scene text detection.
As shown in Fig. \ref{fig:illustration}, the network produces initial prediction with direct regression.
With a novel quadrilateral RoI  convolution (QRC) operation tailed for multi-oriented scene text, the initial quadrilateral prediction is encoded into the feature maps which can be further used to predict offset between the initial prediction and the ground-truth as well as output a refined confidence score.
The refined prediction is obtained with the initial prediction and the predicted offset.
The design of the whole network follows the spirit of anchor-free regression in a fully convolutional manner.
With corner refinement and scoring, the proposed text detector can detect long texts and decline low-quality detections produced in initial prediction.

The contributions of this work could be summarized as follows:
\begin{itemize}
\item We propose a novel build-in module to encode an initial quadrilateral prediction into the feature maps with a light deformable convolution operation.
\item We embed the proposed module into the network to perform corner refinement and scoring in the baseline detector, leading to a fully convolutional corner refinement network for multi-oriented scene text detection.
\item Experimental results on several datasets show our method can outperform the state-of-the-art methods.
The ablation study shows the effectiveness of the corner refinement and scoring for accurate localization.

\end{itemize}

%-------------------------------------------------------------------------
\section{Related Work}

As a basic component of OCR, scene text detection has been a hot research topic for a long time.
With the powerful ability of feature representation and characteristics of end-to-end optimization, recent works on scene text detection are almost based on deep learning.
We roughly review recent works by dividing them into anchor-based methods and anchor-free methods, then we review works on adaptive convolutional sampling.

\noindent\textbf{Anchor-based methods}
\cite{liu2017deep,liao2018textboxes++,ma2018arbitrary,liao2018rotation,tian2016detecting,shi2017detecting,he2017single,yang2018inceptext,xie2019scene,qin2019curved} assume a set of prior boxes and simplify the learning problem to learn relative offsets to anchors.
Recent text detection methods benefit a lot from classic generic object detection methods.
DMPNet \cite{liu2017deep}, SegLink \cite{shi2017detecting}, SSTD \cite{he2017single}, Textboxes++ \cite{liao2018textboxes++} and RRD \cite{liao2018rotation} are based on SSD \cite{liu2016ssd}.
CTPN \cite{tian2016detecting} and RRPN \cite{ma2018arbitrary} are based on Faster R-CNN \cite{ren2015faster}, and SPCNet \cite{xie2019scene} is based on Mask R-CNN \cite{he2017mask}.
Anchors of different scales and aspect ratios are used to model the large variations in scales and aspect ratios of scene text.
However, different datasets usually require different settings to achieve the best performance, which makes it inflexible to transfer between different datasets.
% RRD \cite{liao2018rotation} uses orientation-sensitive features to model the large variations of rotations and an inception block is used to enlarge the receptive field of the detection heads.
% In \cite{ma2018arbitrary}, RRPN and RRoI pooling are proposed to deal with rotations in scene text detection corresponding to RPN and RoI pooling in Faster R-CNN.

% ITN\cite{wang2018geometry} predicts affine transformation parameters for each instance,
% which can not deal with severe perspective distortion (corresponding to perspective transformation) text well.
% Moreover, it is not directly target to detection and cumulative error from the transformation prediction will degrade the overall performance.
% In contrast, In FC2RN, quadrilateral RoI convolution is used to encode initial quadrilateral prediction into the features and
% representation is directly obtained from initial prediction and can deal with perspective distortion.

\noindent\textbf{Anchor-free methods} either directly predict geometry of scene text \cite{zhou2017east,he2017deep,zhang2019look,wang2018geometry,long2018textsnake}
or segmentation confidence score \cite{li2018shape,Zhang_2016_CVPR,deng2018pixellink,tian2019learning,wang2019efficient,db} of being text or not.
In EAST \cite{zhou2017east}, a clear and effective pipeline is proposed to regress text geometry representation of rotated bounding boxes or quadrilaterals.
DeepReg \cite{he2017deep} also directly regresses the four corners of quadrilaterals for multi-oriented text detection.
TextSnake \cite{long2018textsnake} performs local geometry regression and then reconstructs text instances, which is able to detect texts of arbitrary shape.
LOMO \cite{zhang2019look} regresses text geometry in a coarse-to-fine manner.
% ITN \cite{wang2018geometry} applies pixel-level affine transformation estimation for multi-oriented scene text instances. However, it falls short in modeling text with perspective transformation.
% Moreover, it is not directly targeting to detection and cumulative error from the transformation prediction will degrade the overall performance.
% In contrast, FC$^2$RN can refine prediction of general quadrilateral and deal with perspective distortion.
Methods based on segmentation view text detection as an instance segmentation problem.
Apart from text score, shrunk kernel score \cite{li2018shape,long2018textsnake,wang2019efficient,db}, link score with neighbors \cite{deng2018pixellink}, or similarity vector \cite{tian2019learning,wang2019efficient} are used to distinguish different text instances.

\noindent\textbf{Adaptive convolutional sampling.} Deformable Convolution \cite{dai2017deformable} introduces learnable offsets in convolutional sampling.
For scene text detection, ITN \cite{wang2018geometry} applies affine transformation estimation for each location.
The predicted affine transformation is then used to deform the sampling grid in regular convolution, leading to scale and orientation robust text detection.
However, the affine transformation modeling falls when dealing with text with perspective transformation.
Moreover, it is not directly targeting to detection and cumulative error from the transformation prediction will degrade the overall performance.
Adaptive Convolution is proposed in Cascade RPN \cite{vu2019cascade} for feature alignment in which the offsets in deformable convolution is determined by the anchors.
Nevertheless, this modeling does not fit for multi-oriented scene text which is tightly bounded by long quadrilaterals.
Inspired by these methods, we propose QRC which is tailed for multi-oriented scene text detection and is able to deal with proposals of arbitrary convex quadrilaterals.
With QRC, the initial geometry prediction is encoded in the features thus the receptive field is changed correspondingly.

%-------------------------------------------------------------------------
\section{Methodology}

\begin{figure*}[!htb]
\begin{center}
   \includegraphics[width=1\linewidth]{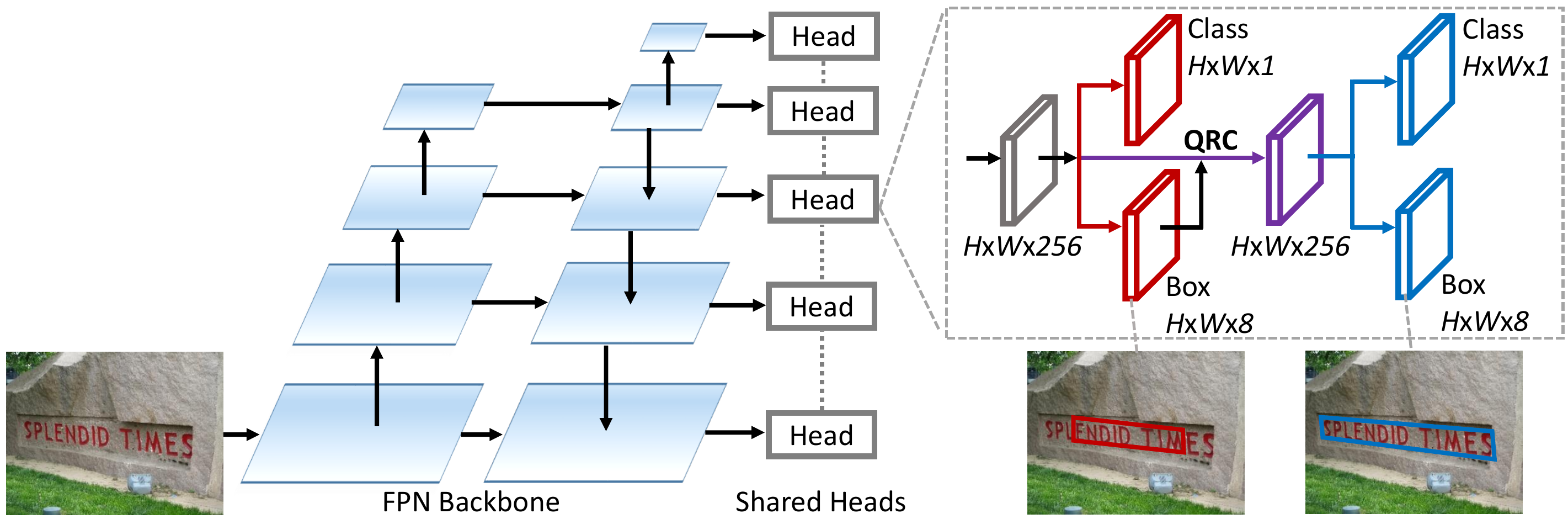}
\end{center}
   \caption{The overall framework of FC$^2$RN. It consists of FPN backbone and shared heads between feature pyramid levels. The red and the blue lines represents the initial prediction and the refined prediction.}
\label{fig:structure}
\end{figure*}

\noindent In this section, we describe the framework of FC$^2$RN in detail.
The baseline model is first introduced.
Next, we introduce QRC.
Then, we describe the corner refinement and scoring heads.
Finally, the learning target and the loss function in optimization are introduced.

%-------------------------------------------------------------------------
\subsection{Baseline Model}

The architecture of FC$^2$RN is illustrated in Fig. \ref{fig:structure}.
We first introduce the baseline model.
We adopt ResNet50 \cite{he2016deep} as the backbone.
Standard FPN \cite{lin2017feature} is used to deal with scale variations of scene text.
Instead of using one level output as those in \cite{zhou2017east,li2018shape,zhang2019look}, we use features from different levels that have different receptive fields naturally to predict texts of different scales.
Pyramid levels of $\{P_2, P_3, P_4, P_5, P_6\}$ are used. The resolution of $P_l$ is $1/2^l$ of the input image.
We use $P_2$ as the first level for better detecting tiny text. The output features of all pyramids have $C = 256$ channels.
We add three consecutive convolutions to the detection heads for further feature extraction.
After that, two $3 \times 3$ convolutions are added to predict the offset to the center point and the classification score respectively.

In training, we project texts of different scales to different pyramid levels.
Instead of using the short side of text as scale measurement, we compute the lengths of the two lines connecting the midpoints of the opposing edge pairs.
The shorter one is taken as the scale of the text instance. This may be a more feasible and robust measurement to describe the scale of texts.
Specifically, texts with scales of $(0, 32)$, $(16, 64)$, $(32, 128)$, $(64, 256)$, $(128, \infty)$ are projected to $P_2$ to $P_6$ respectively.

Given $\bm{\mathrm{x}} \in \mathbb{R}^{H\times W \times C}$ as one level of the feature maps and $s$ as the stride before the layer, for location $\bm{\mathrm{p}} = (p_x, p_y)$ on feature map $\bm{\mathrm{x}}$, the center of the feature bin can be computed as $(\lfloor \frac{s}{2} \rfloor + p_{x}s, \lfloor \frac{s}{2} \rfloor + p_{y}s)$.
The ground-truth quadrilaterals are from top-left to bottom-left in clockwise order.
The learning target is the offsets between the center of the feature bin and the corresponding four corners of the ground-truth, normalized by the stride $s$.
The quadrilateral prediction can be obtained by the center coordinate and the offset prediction $\bm{\mathrm{o}}_i \in \mathbb{R}^{H\times W \times 8}$.

%-------------------------------------------------------------------------
\subsection{Quadrilateral RoI Convolution}
\label{QRC}

\begin{figure}[!htb]
\begin{center}
   \includegraphics[width=0.6\linewidth]{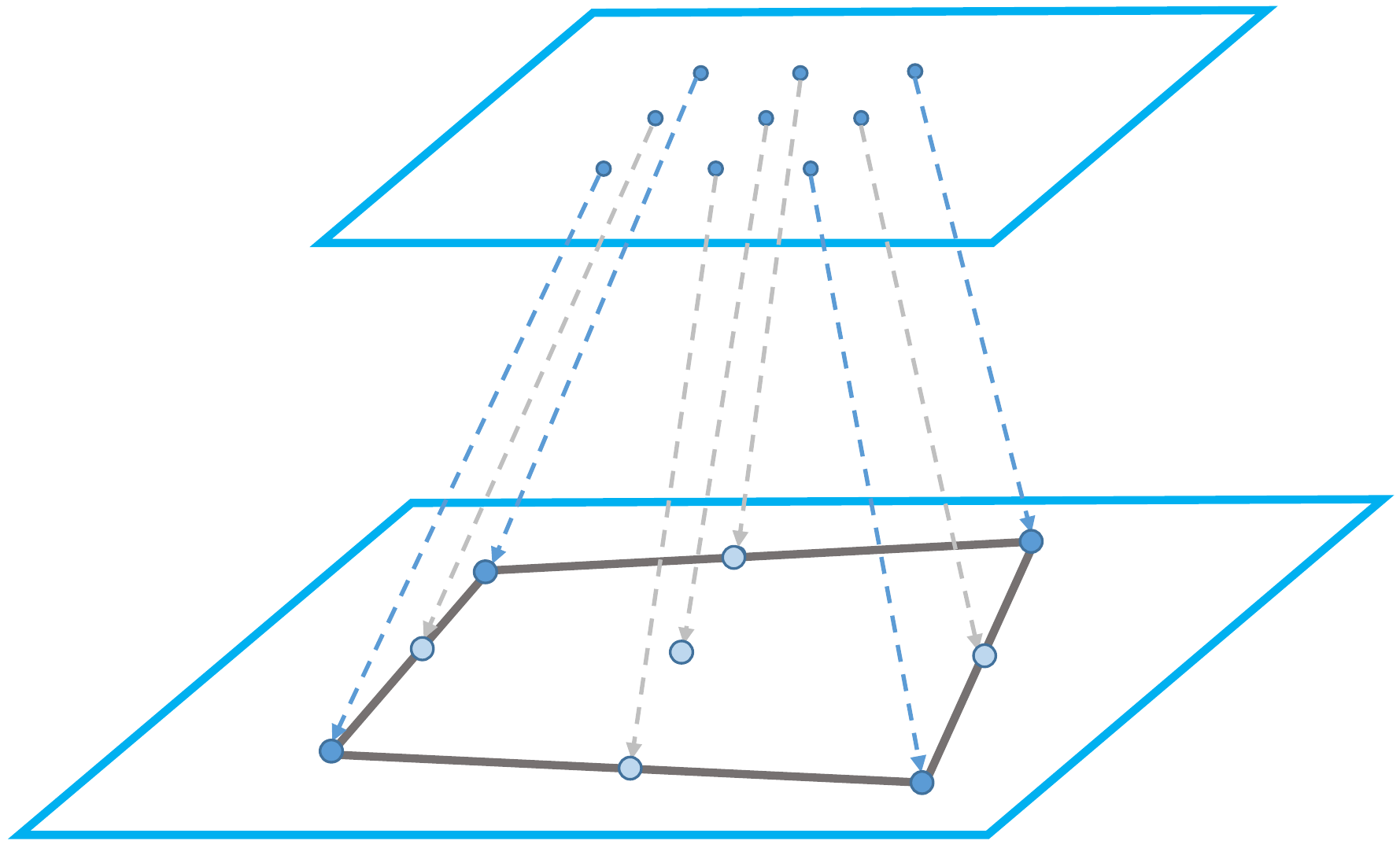}
\end{center}
   \caption{Illustration of $3 \times 3$ QRC. The gray quadrilateral represents quadrilateral prediction. The solid points on the quadrilateral are corners and the light blue points are obtained by uniformly sampling.
   }
\label{fig:sampling}
\end{figure}

Given a feature map $\bm{\mathrm{x}}$, in standard 2D convolution, the feature map is first sampled using a regular grid $\mathcal R = \{(r_x, r_y)\}$, and the samples are summed up with the weight $\bm{\mathrm{w}}$ whose kernel size is $h \times w$.
Here, the grid $\mathcal R$ is defined by the kernel size and dilation.
For each location $\bm{\mathrm{p}}$ on the output feature map $\bm{\mathrm{y}}$, we have:
\begin{equation}
\bm{\mathrm{y}}(\bm{\mathrm{p}}) = \sum_{\bm{\mathrm{r}} \in \mathcal R} \bm{\mathrm{w}}(\bm{\mathrm{r}}) \cdot \bm{\mathrm{x}}(\bm{\mathrm{p}}+\bm{\mathrm{r}}).
\end{equation}
In QRC, the regular gird R is augmented with offsets $\mathrm{\Delta} \bm{\mathrm{r}}$ which is inferred from the new grid $\mathcal G(\bm{\mathrm{p}}) = \{\bm{\mathrm{g}}_{i, j}(\bm{\mathrm{p}})| i =0, 1, ..., h-1;j = 0, 1, ..., w-1 \}$ that is generated by uniformly sampling on the initial quadrilateral prediction $\bm{\mathrm{q}}$ based on $\bm{\mathrm{x}}$.
\begin{equation}
\label{eq:QRC}
\bm{\mathrm{y}}(\bm{\mathrm{p}}) = \sum_{\bm{\mathrm{p}} + \bm{\mathrm{r}} + \mathrm{\Delta} \bm{\mathrm{r}} \in \mathcal G(\bm{\mathrm{p}})}
\bm{\mathrm{w}}(\bm{\mathrm{r}}) \cdot \bm{\mathrm{x}} (\bm{\mathrm{p}} + \bm{\mathrm{r}} + \mathrm{\Delta} \bm{\mathrm{r}}).
\end{equation}
Let ($\bar{\bm{\mathrm{q}}}_1$, $\bar{\bm{\mathrm{q}}}_2$, $\bar{\bm{\mathrm{q}}}_3$, $\bar{\bm{\mathrm{q}}}_4$) denote the projection (corners from top left to bottom left) of $\bm{\mathrm{q}}$ onto the feature map, each position in the grid is obtained by
\begin{equation}
\bm{\mathrm{g}}_{i, j} = K_i^h (K_j^w(\bar{\bm{\mathrm{q}}}_1, \bar{\bm{\mathrm{q}}}_2) + K_j^w(\bar{\bm{\mathrm{q}}}_4, \bar{\bm{\mathrm{q}}}_3)),
\end{equation}
where $K_\cdot^\cdot(\cdot, \cdot)$ corresponds to a linear kernel and can be computed as follows:
\begin{equation}
K_a^b(\bm{\mathrm{m}}, \bm{\mathrm{n}}) = \frac{b - 1- a}{b - 1} \cdot \bm{\mathrm{m}} +  \frac{a}{b - 1} \cdot \bm{\mathrm{n}},
\end{equation}
where $a$, $b$, $\bm{\mathrm{m}}$, $\bm{\mathrm{n}}$ correspond to kernel index, kernel size and two corner points.
Obviously, $\mathcal G(\bm{\mathrm{p}})$ is a linear combination of the corresponding four corners of the quadrilateral prediction.
For example, the nine sampling positions of the $3\times3$ QRC are the midpoints of four edges, the center of the quadrilateral and the four corners as illustrated in Fig.\ref{fig:sampling}.
Given Eq.\ref{eq:QRC}, QRC can be easily implemented with a deformable convolution layer \cite{dai2017deformable}.
It is worth noting that the proposed QRC requires no additional computation compared with the vanilla convolution, which enables it to be integrated into any existing regression-based multi-oriented text detectors seamlessly.

\vspace{-30px}

%-------------------------------------------------------------------------
\subsection{Corner Refinement and Scoring}

\begin{figure}[!htb]
\begin{center}
   \includegraphics[width=0.6\linewidth]{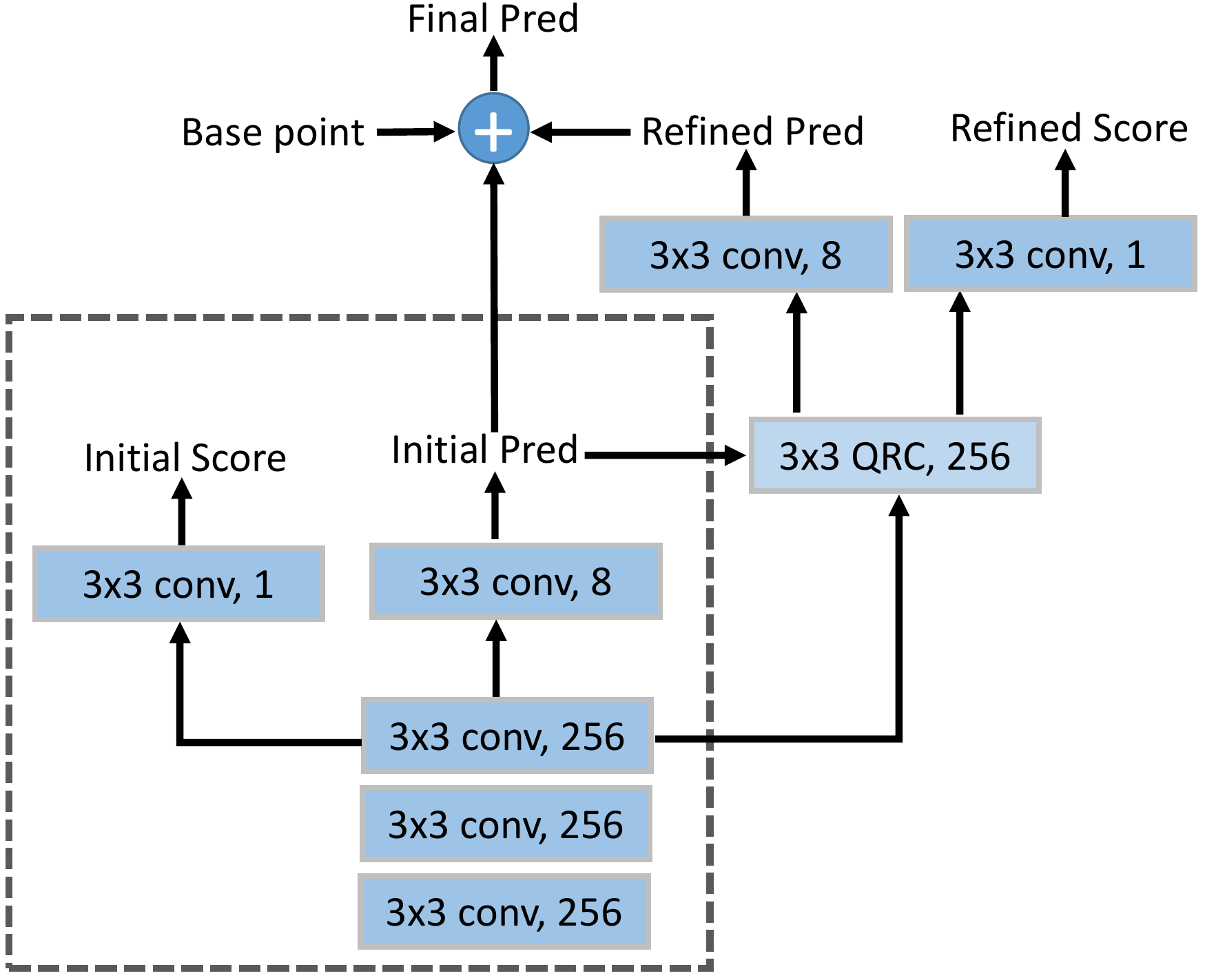}
\end{center}
   \caption{The head architecture of FC$^2$RN. The part in the dashed gray box is the head architecture of the baseline model.}
\label{fig:head}
\end{figure}

The architecture of detection head is illustrated in Fig. \ref{fig:head}.
The initial prediction and score are produced based on the initial feature maps.
We use initial features and initial prediction to produce the refined features as described in Sec. \ref{QRC}.
Then two $3 \times 3$ convolution heads are added to perform corner refinement and scoring. With QRC, the initial quadrilateral prediction is encoded into the initial features which act as attention mechanism that focuses on text instances rather than backgrounds.
With initial prediction encoded, the refinement head is able to predict offsets between the initial prediction and the ground-truth, which can produce more accurate localization results. The scoring head predicts a new classification score on the refined features which has a more appropriate receptive field and is more discriminative in distinguishing foregrounds and backgrounds.

%-------------------------------------------------------------------------
\subsection{Optimization}

Due to the fully convolutional nature of the proposed network, the pipeline of the network is quite simple and can be optimized end-to-end.
Given one feature bin on the output feature maps, corresponding initial score, initial quadrilateral prediction, refined score and refined quadrilateral prediction are obtained at one pass.

\par\noindent\textbf{Label generation}.
We use the rules in \cite{liao2018textboxes++} to decide the order of four corners.
The positive samples of initial classification and regression tasks are the shrunk version of text regions.
Instead of shrinking four edges with the same distance, we shrink four edges with the same proportion.
We argue that feature bins near edges may be not appropriate to regress the whole text when dealing with long texts and tend to produce local incomplete predictions.
Moreover, feature bins near the edge region are more likely to be outliers which may dominate the gradient during training.
If the center of the feature bin falls into the shrunk version of a specific ground-truth, this ground-truth is assigned to the feature bin in training.
The negative samples are the feature bins that don't fall into any ground-truth quadrilateral.
Feature bins in do-not-care regions are ignored during training.

Label or learning targets for corner refinement and scoring are as follows.
The two branches predict offsets between the initial prediction and the ground-truth and new classification score based on the refined features.
Ideally, the IoU (intersection over union) between the prediction quadrilateral and the ground-truth is a good criterion for measuring how good the prediction is.
We consider a feature bin as a positive sample if its IoU between a ground-truth is higher than a threshold.
And negative samples are the ones whose IoU between any ground-truth quadrilateral is lower than the threshold.
For each positive feature bin, the learning target is the ground-truth which has the highest IoU with it.
However, the computation for the IoU of two quadrilaterals is hard to parallel.
In practice, we use the IoU between the minimum bounding boxes of the initial prediction and the ground-truth instead.
We set the IoU threshold as 0.5 and find it works well in practice.

\par\noindent\textbf{Loss function.}
The classification losses used in the network are focal loss \cite{lin2017focal}:
\begin{equation}
L_{cls} = -\alpha_{t}(1 - p_t)^{\gamma}log(p_t),
\end{equation}
where
\begin{equation}
p_t =
\begin{cases}
p & \text{y = 1} \\
1 - p & \text{otherwise},
\end{cases}
\end{equation}
$y$ and $p$ are the label and prediction. In both the classification losses, we set $\alpha_t = 0.25$ and $\gamma = 2$.

The regression losses we used are smooth L1 loss \cite{girshick2015fast}:
\begin{equation}
L_{reg} = \frac{1}{8}\sum_{i=1}^{8}smooth_{L_1}(c_i - \hat{c}_i),
\end{equation}
where $c_i$ is i-th coordinate offset between detected quadrilateral and the ground-truth and $\hat{c}_i$ is the corresponding predicted value.
The whole loss function of the network is formulated as follows:
\begin{equation}
L = L_{cls_i} + \lambda_{1}L_{reg_i} + \lambda_{3}(L_{cls_r} + \lambda_{2}L_{reg_r}),
\end{equation}
$L_{cls_i}$, $L_{reg_i}$, $L_{cls_r}$, and $L_{reg_r}$ is the loss for the initial classification, initial regression, refined classification and refined regression.
The balanced factors $\lambda_1$, $\lambda_2$ and $\lambda_3$ are all set to 1.

During inference, the refined quadrilateral prediction can be obtained by
\begin{equation}
\bm{\mathrm{o}}_q = \bm{\mathrm{o}}_c + (\bm{\mathrm{o}}_i + \bm{\mathrm{o}}_r) \times s,
\end{equation}
where $\bm{\mathrm{o}}_c$, $\bm{\mathrm{o}}_i$, $\bm{\mathrm{o}}_r$, $s$ are the center coordinate of the feature bin, the initial offset prediction, the refined offset prediction and the stride, scored by the refined score $\bm{\mathrm{s}}_r$.
%-------------------------------------------------------------------------
\section{Experiments}

In this section, we evaluate our approach on MSRA-TD500 \cite{yao2012detecting}, ICDAR2017-RCTW \cite{shi2017icdar2017}, ICDAR2015 \cite{karatzas2015icdar} and COCO-Text \cite{cocotext} to show the effectiveness of our approach.
%-------------------------------------------------------------------------
\subsection{Datasets}

\par\noindent\textbf{MSRA-TD500} is a multilingual dataset focusing on oriented text lines.
Large variations of text scales and orientations are presented in this dataset.
It consists of 300 training images and 200 testing images.

\par\noindent\textbf{ICDAR2017-RCTW} comprises 8034 training images and 4229 test images with scene texts printed in either Chinese or English.
The images are captured from different sources including street views, posters, screen-shot, etc.
Multi-oriented words and text lines are annotated using quadrangles.

\par\noindent\textbf{ICDAR2015} is a multi-oriented text detection dataset only for English, which includes 1000 training images and 500 testing images.
The text regions are annotated with quadrilaterals.

\par\noindent\textbf{COCO-Text} is a large dataset that contains 63686 images, where 43,686 of the images are used for training,
10,000 for validation, and 10,000 for testing. It is one of the challenges of the ICDAR 2017 robust reading competition.
The dataset is quite challenging due to the diversity of text in natural scenes.

%-------------------------------------------------------------------------
\subsection{Implementation Details}

Our work is implemented based on MMDetection \cite{mmdetection}. ImageNet pre-trained model is used to initialize the backbone.
We use SGD as optimizer with batch size 2, momentum 0.9 and weight decay 0.0001 in training.
The number of maximum iterations in training is 48 epochs. We adopt warm-up in the initial 500 iterations.
The initial learning rate is set to 0.00125 for all experiments, and decayed by 0.1 on the 32nd and 44th epoch.
The shrink factor of the text region in training is set to 0.25.
All the experiments are performed on GeForce GTX 1080 Ti.

Due to the imbalanced distribution of text scales and orientations, we adopt random rotation, random crop, random scale, random flip as data augmentation in training.
% And multi-scale training is also used.
% The resolutions of training images are randomly selected from 680 to 1000 with the interval of 40, and the maximum size is maintained at 1480.
For testing, we only use single scale testing for all datasets because public methods have numerous settings for multi-scale testing, which is hard to give a fair comparison. The scale and maximum size are (1200, 1600).
We adopt polygonal non-maximum suppression (PNMS) proposed in \cite{liu2019curved} to suppress redundant detections.

%-------------------------------------------------------------------------
\subsection{Ablation Study}

We perform ablation study on MSRA-TD500 with different settings to analyze the function of corner refinement and scoring.
To better illustrate the ability to detect long texts, we use 4k well annotated samples from \cite{shi2017icdar2017} for pretraining which is adopted in \cite{liu2019omnidirectional}.
The result is in Table \ref{tab:ablation}. With corner refinement, the F-measure is higher than the baseline model by 4.1 F-measure.
When we replace the initial confidence score with the refined confidence score, we achieve 1.1 more increasing on F-measure.
The result shows the effectiveness of corner refinement and scoring quantitatively.

IoU threshold of 0.5 is usually adopted in detection. However, it is not enough for accurate scene text detection and subsequent text recognition task.
We further constrain the IoU threshold to 0.75 for better illustration. As is shown in Table \ref{tab:ablation}, the F-measure of three methods drops by 31.6, 17.1, 13.4 respectively compared with those under 0.5 IoU metric.
With corner refinement, 18.6 F-measure increment is obtained upon the baseline.
The performance obtains another increment of 4.8 F-measure when using the refined score.
This shows the effectiveness of QRC and the two subtasks clearly under a high IoU threshold.

% \vspace{-15px}
\begin{table}[!htb]
\begin{center}
\caption{Evaluation results on MSRA-TD 500 with different model settings. "Baseline", "CR" and "CS" represent the baseline model, corner refinement and corner scoring respectively. “P”, “R”, and “F” indicate precision, recall, and F-measure respectively.}
\label{tab:ablation}
\setlength{\tabcolsep}{3mm}{
\begin{tabular}{|c|c|c||c|c|c||c|c|c|}
\hline
\multirow{2}{*}{Baseline} & \multirow{2}{*}{\;CR\;} & \multirow{2}{*}{\;CS\;} &
   \multicolumn{3}{|c||}{IoU@0.5} & \multicolumn{3}{|c|}{IoU@0.75} \\
   \cline{4-9}
   &&&P & R & F &P & R & F \\
\hline\hline
$\surd$ & $\times$ & $\times$ & 82.8 & 82.1 & 82.5 & 51.1& 50.7 & 50.9 \\
\hline
$\surd$ & $\surd$ & $\times$ & 89.5 & 83.8 & 86.6 & 71.4 & 67.7 & 69.5 \\
\hline
$\surd$ & $\surd$ & $\surd$ & {\bf 90.3} & {\bf 85.2} & {\bf 87.7} & {\bf 80.5} & {\bf 69.0} & {\bf 74.3} \\
\hline
\end{tabular}}
\end{center}
\end{table}
% \vspace{-15px}

\begin{figure}[htbp]
\centering
\subfigure[Baseline]{
\begin{minipage}[t]{0.23\linewidth}
\centering
\includegraphics[width=1.1in]{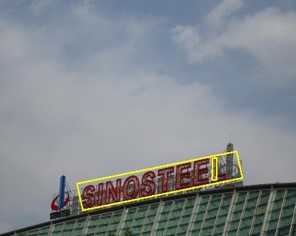}
\includegraphics[width=1.1in]{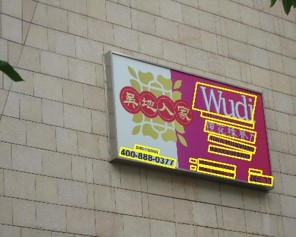}
\includegraphics[width=1.1in]{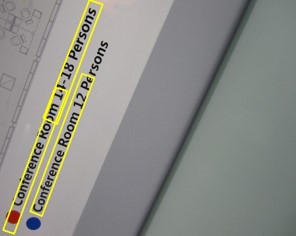}
\end{minipage}
}
\subfigure[Baseline + CR]{
\begin{minipage}[t]{0.23\linewidth}
\centering
\includegraphics[width=1.1in]{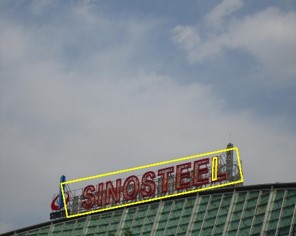}
\includegraphics[width=1.1in]{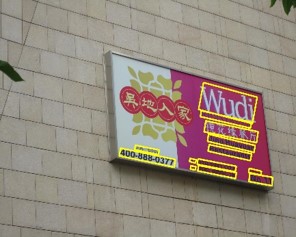}
\includegraphics[width=1.1in]{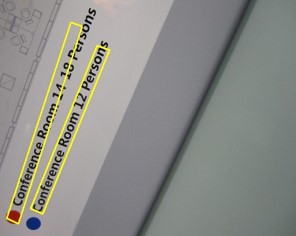}
\end{minipage}
}
\subfigure[Baseline + CR + CS]{
\begin{minipage}[t]{0.23\linewidth}
\centering
\includegraphics[width=1.1in]{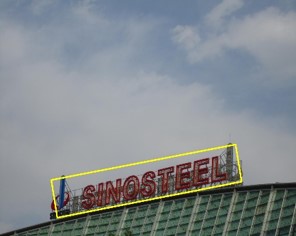}
\includegraphics[width=1.1in]{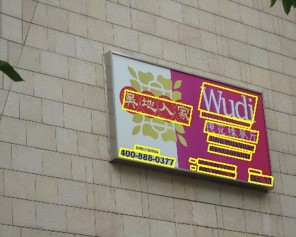}
\includegraphics[width=1.1in]{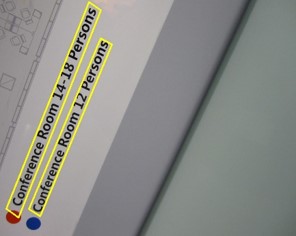}
\end{minipage}
}
\subfigure[GT]{
\begin{minipage}[t]{0.23\linewidth}
\centering
\includegraphics[width=1.1in]{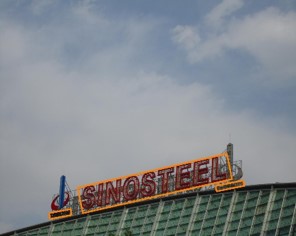}
\includegraphics[width=1.1in]{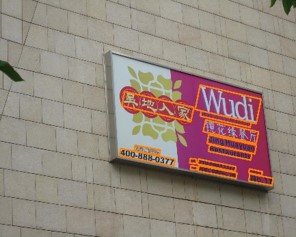}
\includegraphics[width=1.1in]{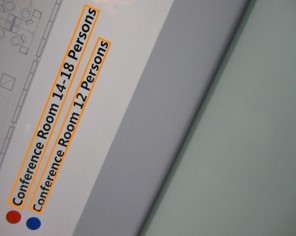}
\end{minipage}
}
\centering
\caption{Visualization of results with different model settings. "CR" and "CS" denote corner refinement and corner scoring.
(a-d) correspond to the detection results of baseline model, baseline model with corner refinement, FC$^2$RN and the ground-truth respectively.
Detection results and the ground-truth are marked with yellow and orange boxes.}
\label{fig:ablation}
\end{figure}

Detection results of three methods and ground-truth are visualized in Fig. \ref{fig:ablation}, which illustrates the function of corner refinement and scoring well.
When detecting long texts, the baseline tends to produce incomplete detections due to the limitation of the receptive field, which severely degrades the performance.
With corner refinement, the model is able to produce more accurate regression results compared with the baseline model.
However, the initial score describes how suitable the feature bin is for direct regression and is not fit for the refined regression.
As a result, more accurate regression results may be suppressed due to the unreasonable scoring process.
The refined score is predicted based on the refined features with initial prediction encoded and measures how well the initial prediction is.
With the refined score, the confidences of detection results are much more reasonable.
Moreover, the refined score also shows stronger ability distinguishing scene text with backgrounds.

% \begin{figure}[!htb]
% \begin{center}
%    \includegraphics[width=0.9\linewidth]{figures/ab.pdf}
% \end{center}
%    \caption{Visualization of results with different model settings. The four columns correspond to the detection results of baseline model, baseline model with corner refinement, FC$^2$RN and the ground-truth respectively. Detection results and the ground-truth are marked with yellow and orange boxes.}
% \label{fig:ablation}
% \end{figure}

%-------------------------------------------------------------------------
\subsection{Comparison with State-of-the-Arts}

\subsubsection{Detecting Oriented Multi-Lingual Text Lines.} We evaluate our method on challenging oriented text line datasets MSRA-TD500 and ICDAR2017-RCTW.

\begin{table}[!htb]
\begin{center}
\caption{Experimental results on MSRA-TD500 and ICDAR2015. “P”, “R”, and “F” indicate precision, recall, and F-measure respectively. "*" denotes end-to-end recognition methods.}
\label{tab:td500+ic15}
\setlength{\tabcolsep}{3mm}{
\begin{tabular}{|c||c|c|c||c|c|c|}
\hline
\multirow{2}{*}{Method} &
   \multicolumn{3}{|c||}{MSRA-TD500} & \multicolumn{3}{|c|}{ICDAR2015} \\
   \cline{2-7}
   & P & R & F & P & R & F \\
\hline\hline
EAST \cite{zhou2017east} & 87.3 & 67.4 & 76.1 & 83.6 & 73.5 & 78.2 \\
SegLink \cite{shi2017detecting} & 86.0 & 70.0 & 77.0 & 73.1 & 76.8 & 75.0 \\
DeepReg \cite{he2017deep} & 77.0 & 70.0 & 74.0 & 82.0 & 80.0 & 81.0 \\
SSTD \cite{he2017single} & - & - & - & 80.2 & 73.9 & 76.9 \\
WordSup \cite{hu2017wordsup} & - & - & - & 79.3 & 77.0 & 78.2 \\
RRPN \cite{ma2018arbitrary} & 82.0 & 68.0 & 74.0 & 82.0 & 73.0 & 77.0 \\
PixelLink \cite{deng2018pixellink} & 83.0 & 73.2 & 77.8 & 85.5 & 82.0 & 83.7 \\
Lyu et al. \cite{lyu2018multi} & 87.6 & 76.2 & 81.5 & {\bf 94.1} & 70.7 & 80.7 \\
RRD \cite{liao2018rotation} & 87.0 & 73.0 & 79.0 & 85.6 & 79.0 & 82.2 \\
MCN \cite{liu2018learning} & 88.0 & 79.0 & 83.0 & 72.0 & 80.0 & 76.0 \\
ITN \cite{wang2018geometry} & 90.3 & 72.3 & 80.3 & 85.7 & 74.1 & 79.5 \\
FTSN \cite{dai2018fused} & 87.6 & 77.1 & 82.0 & 88.6 & 80.0 & 84.1 \\
IncepText \cite{yang2018inceptext} & 87.5 & 79.0 & 83.0 & 90.5 & 90.6 & 85.3 \\
TextSnake \cite{long2018textsnake} & 83.2 & 73.9 & 78.3 & 84.9 & 80.4 & 82.6 \\
Border \cite{xue2018accurate} & 83.0 & 77.4 & 80.1 & - & - & - \\
TextField \cite{xu2019textfield} & 87.4 & 75.9 & 81.3 & 84.3 & 80.5 & 82.4 \\
SPCNet \cite{xie2019scene} & - & - & - & 88.7 & 85.8 & 87.2 \\
PSENet-1s \cite{li2018shape} & - & - & - & 86.9 & 84.5 & 85.7 \\
CRAFT \cite{baek2019character} & 88.2 & 78.2 & 82.9 & 89.8 & 84.3 & 86.9 \\
SAE \cite{tian2019learning} & 84.2 & 81.7 & 82.9 & 88.3 & 85.0 & 86.6 \\
Wang et al. \cite{wang2019arbitrary} & 85.2 & 82.1 & 83.6 & 89.2 & 86.0 & 87.6 \\
LOMO \cite{zhang2019look} & - & - & - & 91.3 & 83.5 & 87.2 \\
MSR \cite{xue2019msr} & 87.4 & 76.7 & 81.7 & 86.6 & 78.4 & 82.3 \\
BDN \cite{liu2019omnidirectional} & 89.6 & 80.5 & 84.8 & 89.4 & 83.8 & 86.5 \\
PAN \cite{wang2019efficient} & 84.4 & {\bf 83.8} & 84.1 & 84.0 & 81.9 & 82.8 \\
GNNets \cite{xu2019geometry} & - & - & - & 90.4 & 86.7 & 88.5 \\
DB \cite{db} & {\bf 91.5} & 79.2 & 84.9 & 91.8 & 83.2 & 87.3 \\
\hline
FOTS* \cite{liu2018fots} & - & - & - & 91.0 & 85.2 & 88.0 \\
Mask TextSpotter* \cite{masktextspotter} & - & - & - & 86.6 & 87.3 & 87.0 \\
CharNet R-50* \cite{xing2019convolutional} & - & - & - & 91.2 & 88.3 & {\bf 89.7} \\
Qin et al.* \cite{qin2019towards} & - & - & - & 89.4 & 85.8 & 87.5 \\
TextDragon* \cite{feng2019textdragon} & - & - & - & 92.3 & 83.8 & 87.9 \\
Boundary* \cite{wang2019all} & - & - & - & 89.8 & 87.5 & 88.6 \\
Text Perceptron* \cite{qiao2020text} & - & - & - & 92.3 & 82.5 & 87.1 \\
\hline\hline
FC$^2$RN & 90.3 & 81.8 & {\bf 85.8} & 89.0 & {\bf 88.7} & {\bf 88.9} \\
\hline
\end{tabular}}
\end{center}
\end{table}

{\bf MSRA-TD500.} Because the training set is rather small, we follow the previous works \cite{zhou2017east,lyu2018multi,long2018textsnake} to include the 400 images from HUST-TR400 \cite{yao2014unified} as training data. The detection results are listed in Table \ref{tab:td500+ic15}.
FC$^2$RN achieves the state-of-the-art performance in terms of F-measure.
The F-measure outperforms the state-of-the-art methods by 0.9 percents.
FC$^2$RN outperforms anchor-free regression method EAST \cite{zhou2017east} and DeepReg \cite{he2017deep} because of the enlarged receptive field and the refinement heads, which enables the network to detect oriented texts with large aspect ratios.
Compared with methods that group or cluster local results to produce final results like PixelLink \cite{deng2018pixellink}, SegLink \cite{shi2017detecting}, Lyu et al. \cite{lyu2018multi}, MCN \cite{liu2018learning} and TextSnake \cite{long2018textsnake}, FC$^2$RN outputs refined results at one pass, which eliminates the cumulative error of intermediate process.
For two-stage Mask R-CNN based methods like FTSN \cite{dai2018fused}, IncepText \cite{yang2018inceptext} and BDN \cite{liu2019omnidirectional}, IoU match criterion using square anchors or proposals may cause confusion in learning because different inclined tiled texts may have nearly overlapping bounding boxes.
Moreover, there may be several inclined tiled long texts in the same bounding box.
This makes the mask branch that performs foreground/background binary classification hard to distinguish which one is the foreground instance and degrades the performance on multi-oriented text.

{\bf ICDAR2017-RCTW.}
No extra data but official RCTW training samples are used in training.
Our single-scale testing results outperform the exiting best single-scale results \cite{yang2018inceptext} for 3.4 F-measure which is a large margin.
And the results also outperform the SOTA method LOMO \cite{zhang2019look} which uses multi-scale testing, which illustrates the ability of the proposed method to deal with challenging long texts.

\begin{table}[!htb]
\begin{center}
\caption{Experimental results on RCTW benchmark. "MS" denotes multi-scale testing.}
\label{tab:rctw}
\setlength{\tabcolsep}{3mm}{
\begin{tabular}{|c||c|c|c|}
\hline
Method & Precision & Recall & F-measure \\
\hline\hline
Official baseline \cite{shi2017icdar2017} & 76.0 & 40.4 & 52.8 \\
EAST \cite{zhou2017east} & 59.7 & 47.8 & 53.1 \\
RRD \cite{liao2018rotation} & 72.4 & 45.3 & 55.7 \\
IncepText \cite{yang2018inceptext} & 78.5 & 56.9 & 66.0 \\
LOMO \cite{zhang2019look} & {\bf 80.4} & 50.8 & 62.3 \\
\hline
RRD MS \cite{liao2018rotation} & 77.5 & 59.1 & 67.0 \\
Border MS \cite{xue2018accurate} & 78.2 & 58.8 & 67.1 \\
LOMO MS \cite{zhang2019look} & 79.1 & 60.2 & 68.4 \\
\hline\hline
FC$^2$RN & 77.5 & {\bf 63.0} & {\bf 69.4} \\
\hline
\end{tabular}}
\end{center}
\end{table}
\vspace{-30px}

%-------------------------------------------------------------------------

\subsubsection{Detecting Oriented English Words.}
To further demonstrate the effectiveness of our method in detecting English words, we evaluate our method on ICDAR2015 and COCO-Text.

{\bf ICDAR2015.}
To compare with the state-of-the-art methods, we follow \cite{li2018shape,xu2019geometry,xie2019scene,liu2018fots} to use ICDAR2017-MLT \cite{nayef2017icdar2017} to pretrain the model.
The pretrained model is finefuned another 48 epochs using ICDAR2015 training data.
The evaluation protocol is based on \cite{karatzas2015icdar}.
As shown in Table \ref{tab:td500+ic15}, FC$^2$RN outperforms all other methods with single scale test performance of 88.9 F-measure.
The recall rate of 88.7 outperforms the existing state-of-the-art method GNNets by nearly 2 percents.
Note that GNNets uses ResNet152 as its backbone whereas our backbone is ResNet50.
Compared with anchor-free detectors, FC$^2$RN obtains 10 percents F-measure increment with EAST \cite{zhou2017east} and outperforms recently proposed methods
LOMO \cite{zhang2019look} and PSENet \cite{li2018shape}.
FC$^2$RN also outperforms other two-stage detectors like IncepText \cite{yang2018inceptext} and SPCNet \cite{xie2019scene}.
Compared with end-to-end recognition which utilizes recognition supervision,
FC$^2$RN outperforms most of the methods except CharNet \cite{xing2019convolutional} in which character-level supervision supervision is also used.

\begin{figure}[!htb]
\centering
\subfigure[MSRA-TD500]{
\begin{minipage}[t]{0.23\linewidth}
\centering
\includegraphics[width=1.1in]{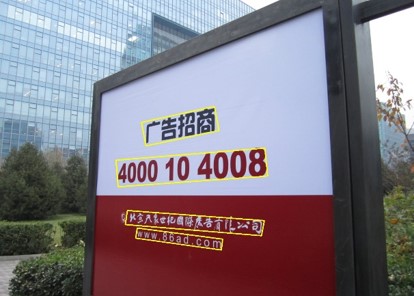}
\includegraphics[width=1.1in]{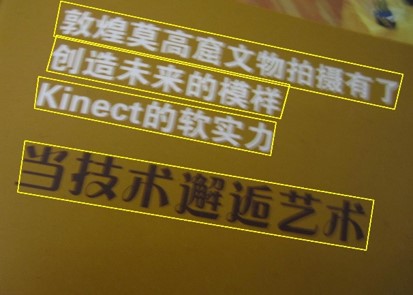}
\includegraphics[width=1.1in]{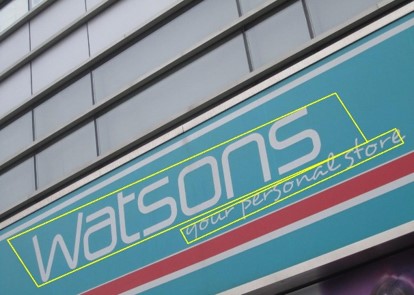}
\end{minipage}
}
\subfigure[RCTW]{
\begin{minipage}[t]{0.23\linewidth}
\centering
\includegraphics[width=1.1in]{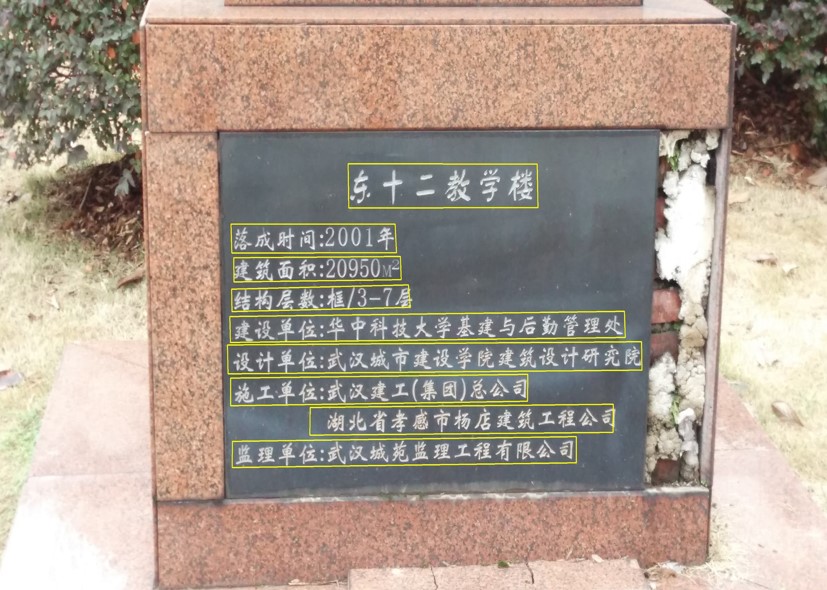}
\includegraphics[width=1.1in]{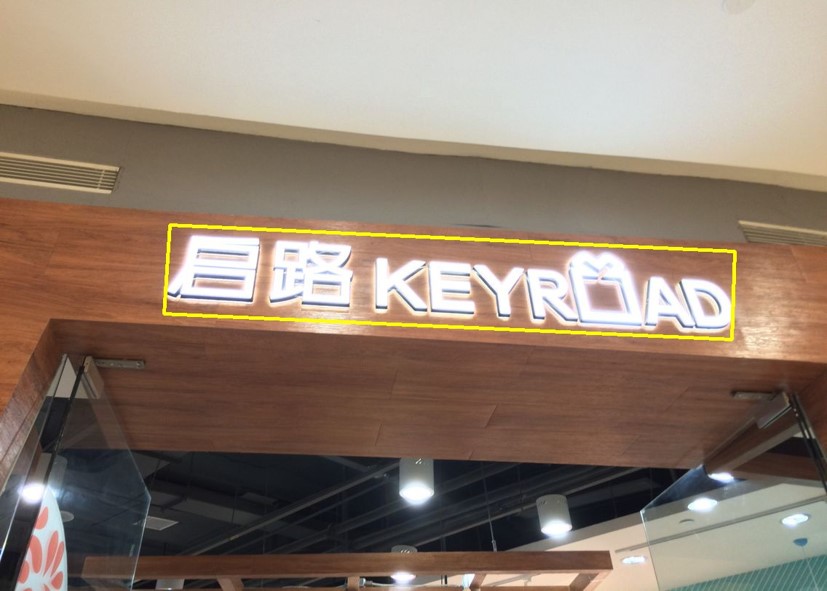}
\includegraphics[width=1.1in]{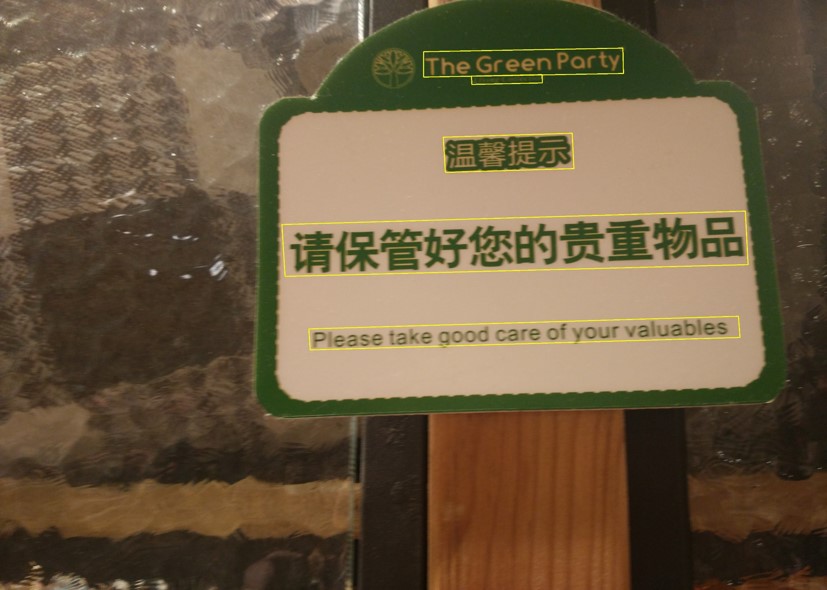}
\end{minipage}
}
\subfigure[ICDAR2015]{
\begin{minipage}[t]{0.23\linewidth}
\centering
\includegraphics[width=1.1in]{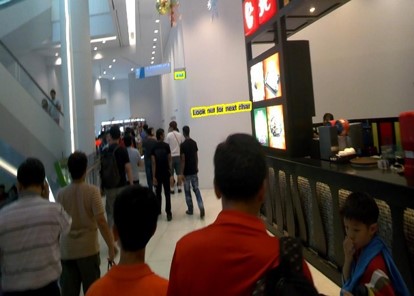}
\includegraphics[width=1.1in]{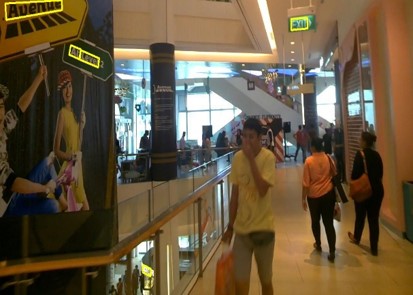}
\includegraphics[width=1.1in]{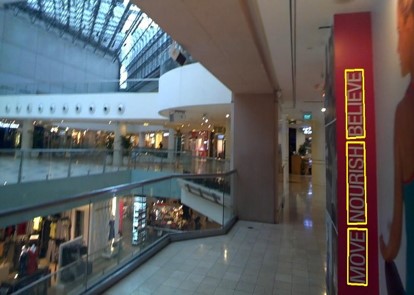}
\end{minipage}
}
\subfigure[COCO-Text]{
\begin{minipage}[t]{0.23\linewidth}
\centering
\includegraphics[width=1.1in]{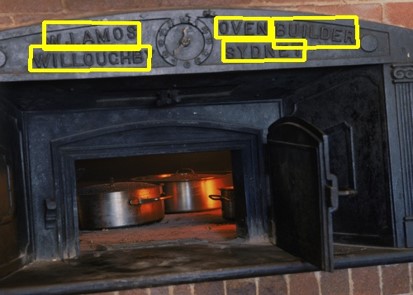}
\includegraphics[width=1.1in]{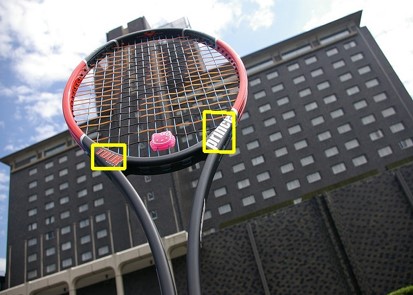}
\includegraphics[width=1.1in]{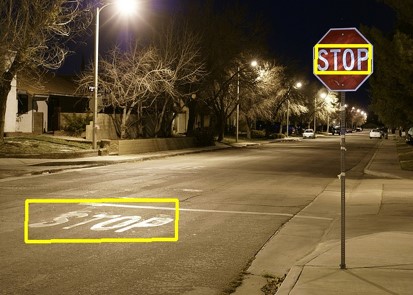}
\end{minipage}
}
\centering
\caption{Examples of detection results. The four columns are results on MSRA-TD500, ICDAR2017-RCTW, ICDAR2015 and COCO-Text respectively.}
\label{fig:results}
\end{figure}

%-------------------------------------------------------------------------
{\bf COCO-Text.}
We evaluate our model on the ICDAR2017 robust reading challenge on COCO-Text \cite{coco-text-chanllenge} with the annotations V1.4.
The results are reported in Table \ref{tab:cocotext}. FC$^2$RN achieves the state-of-the-art performance among all the methods.
Our single scale result outperforms RRD \cite{liao2018rotation} and Lyu at el. \cite{lyu2018multi} which use multi-scale test.
FC$^2$RN also outperforms end-to-end recognition method Mask TextSpotter \cite{masktextspotter} which uses character-level annotations in training.
FC$^2$RN is also comparable to the newly proposed end-to-end recognition method Boundary \cite{wang2019all}.
Moreover, under the IoU metric of 0.75, FC$^2$RN achieves better performance than other methods, in which 1.3 F-measure increment is obtained than the state-of-the-art method RRD with multi-scale testing.
Compared with Lyu et al. MS, FC$^2$RN obtains more improvement at IoU 0.75 than that at IoU 0.5, which reveals that FC$^2$RN outputs more accurate quadrilaterals.
Qualitative results are shown in Fig. \ref{fig:results}.

\begin{table}[!htb]
\begin{center}
\caption{Experimental results on COCO-Text Challenge. "MS" denotes multi-scale testing. “P”, “R”, and “F” indicate precision, recall, and F-measure respectively. "*" denotes end-to-end recognition methods.}
\label{tab:cocotext}
\setlength{\tabcolsep}{3mm}{
\begin{tabular}{|c||c|c|c||c|c|c|}
\hline
\multirow{2}{*}{Method} &
   \multicolumn{3}{|c||}{IoU@0.5} & \multicolumn{3}{|c|}{IoU@0.75} \\
   \cline{2-7}
   & P & R & F & P & R & F \\
\hline\hline
UM \cite{coco-text-chanllenge}& 47.6 & 65.5 & 55.1 & 23.0 & 31.0 & 26.0 \\
TDN SJTU v2 \cite{coco-text-chanllenge} & 62.4 & 54.3 & 58.1 & 32.0 & 28.0 & 30.0 \\
Text Detection DL \cite{coco-text-chanllenge} & 60.9 & 61.8 & 61.4 & 38.0 & 34.0 & 36.0 \\
Lyu et al. \cite{lyu2018multi} & {\bf 72.5} & 52.9 & 61.1 & 40.0 & 30.0 & 34.6 \\
\hline
Lyu et al. MS \cite{lyu2018multi} & 62.9 & {\bf 62.2} & 62.6 & 35.1 & {\bf 34.8} & 34.9 \\
RRD MS \cite{liao2018rotation} & 64.0 & 57.0 & 61.0 & 38.0 & 34.0 & 36.0 \\
Mask TextSpotter* \cite{masktextspotter} & 66.8 & 58.3 & 62.3 & - & - & - \\
Boundary* \cite{wang2019all} & 67.7 & 59.0 & {\bf 63.0} & - & - & - \\
\hline\hline
FC$^2$RN & 68.5 & 58.2 & {\bf 63.0} & {\bf 44.7} & 32.1 & {\bf37.3} \\
\hline
\end{tabular}}
\end{center}
\end{table}
\vspace{-30px}

% \begin{figure}[!htb]
% \begin{center}
%    \includegraphics[width=0.9\linewidth]{figures/results.pdf}
% \end{center}
%    \caption{Examples of detection results. The four columns are results on MSRA-TD500, ICDAR2017-RCTW, ICDAR2015 and COCO-Text respectively.}
% \label{fig:results}
% \end{figure}

%-------------------------------------------------------------------------
\subsection{Runtime Analysis}
The speed of FC$^2$RN is compared with the baseline model as shown in Table \ref{tab:speed}. FC$^2$RN is only a little slower than the baseline model.
The additional parameters and computation are from the QRC and refinement heads.
The refinement module adds only 0.61 M parameters on the base model with 28.42 M parameters.
When the input size is $1280 \times 800$, it rises 50.3 GMac more computation than the baseline model with 297.0 GMac.
We find that the computation mainly comes from the dense convolution computation on pyramid level $P_2$ which has $1/4$ resolution of the input image.

% \vspace{-15px}
\begin{table}[!htb]
\begin{center}
\caption{Comparison on speed of different scales, model complexity and computation with $1280\times800$ input between the baseline model and FC$^2$RN.}
\label{tab:speed}
\setlength{\tabcolsep}{4mm}{
\begin{tabular}{|c||c|c|c|c|}
\hline
\multirow{2}{*}{Method} &\multicolumn{2}{|c|}{Scale} & \multirow{2}{*}{Flops} & \multirow{2}{*}{Params}\\
   \cline{2-3}
   & 928 & 736 & & \\
\hline\hline
Baseline & 7.7 FPS & 11.0 FPS & 297.0 GMac & 28.42 M \\
\hline
FC$^2$RN & 5.8 FPS & 8.8 FPS & 347.3 GMac & 29.03 M \\
\hline
\end{tabular}}
\end{center}
\end{table}
% \vspace{-35px}

%-------------------------------------------------------------------------
\subsection{Limitation}
One limitation is that our method can not deal with text lines with large character spacing well since there are few samples in the training set.
This also happens in many other methods.
% Failure cases of FC$^2$RN are shown in Fig.\ref{fig:limitation}.
% One limitation is that it can't handle curve text well since there are few curve samples in the training set.
% Besides, our method fails to detect text lines with large character spacing.

% \vspace{-15px}
% \begin{figure*}[!htb]
% \setlength{\abovecaptionskip}{0.0cm}
% \begin{center}
%    \includegraphics[width=0.6\linewidth]{figures/limitation.pdf}
% \end{center}
%    \caption{Failure cases of our method. The dashed boxes in red are missing ground truths. The green boxes are our predictions.}
% \label{fig:limitation}
% \end{figure*}
% \vspace{-30px}

%-------------------------------------------------------------------------
\section{Conclusion}

In this work, we propose a novel fully convolutional corner refinement network  (FC$^2$RN) for accurate multi-oriented scene text detection.
FC$^2$RN has a simple yet effective pipeline.
Compared with the baseline model, only several convolutions are added.
With the prediction of direct regression encoded, the network is able to refine the initial prediction and produce a new score.
Experiments on several datasets show the effectiveness of our method, especially on long text dataset.
Furthermore, due to the light-weight and fully convolutional nature of our detector,
it can be seamlessly integrated with other frameworks like FOTS \cite{liu2018fots}, TextNet \cite{sun2018textnet} for end-to-end recognition
and LOMO \cite{zhang2019look} for better detecting curve text.
Tricks like iterative refinement \cite{zhang2019look} at the testing time could be also included for better performance but not involved in this work.
In the future, we will extend this work to end-to-end text spotting.

%===========================================================
\bibliographystyle{splncs}
\bibliography{egbib}

%this would normally be the end of your paper, but you may also have an appendix
%within the given limit of number of pages
\end{document}